%% file: ieee_main.tex
\documentclass[letterpaper, 10 pt, conference]{ieeeconf}  

\IEEEoverridecommandlockouts                              

\usepackage{graphicx} 
\usepackage{amsmath} 
\usepackage{amssymb}  
\usepackage{cite}
\usepackage{amsmath,amssymb,amsfonts}
\usepackage{algorithmic}
\usepackage{textcomp}
\usepackage{xcolor}
\usepackage{booktabs}
\usepackage{xspace}
\usepackage{multirow}
\input{include/commands}

\title{\LARGE \bf
HydraCollab: Adaptive Collaborative-Perception for Distributed Autonomous Systems
}

\author{Luke Chen~\and Cheng-Ju Wu~\and David R. Martin$^{*}$~\and Qilin Ye~\and Pramod Khargonekar~\and Mohammad Abdullah Al Faruque
    \thanks{
        *Corresponding Author.
        Department of Electrical Engineering and Computer Science, University of California, Irvine, USA.
        {\tt\small~\{panwangc, cwu30, davidrm3, qiliny3, pramod.khargonekar, alfaruqu\}@uci.edu}.
    }%
}

\begin{document}

\maketitle
\thispagestyle{empty}
\pagestyle{empty}

\input{iros_camera_ready_sections/0_iros_abstract}

\input{iros_camera_ready_sections/1_iros_intro}
\input{iros_camera_ready_sections/2_iros_related}
\input{iros_camera_ready_sections/3_iros_problem}
\input{iros_camera_ready_sections/4_iros_method}
\input{iros_camera_ready_sections/5_iros_exp}
\input{iros_camera_ready_sections/6_iros_conclusion}








\bibliographystyle{IEEEtran}
\bibliography{ref}

\end{document}

%% file: include/commands.tex
\newcommand{\ourmethod}{\textit{HydraCollab}\xspace}

\definecolor{darkgreen}{RGB}{0,120,0}

%% file: iros_camera_ready_sections/0_iros_abstract.tex
\begin{abstract}
Collaborative-perception enables multi-robot systems  to enhance situational awareness by sharing perceptual information. 
Existing collaborative-perception systems face an inherent trade-off between communication bandwidth requirements and perception accuracy, where methods that exchange more information achieve better perception results at the cost of increased communication overhead.
However, real-world communication networks impose bandwidth constraints that require minimizing communication overhead without sacrificing perception performance.
To address this challenge, we propose \ourmethod, an adaptive collaborative-perception framework that (i) selectively transmits the most informative sensor features and (ii) dynamically employs collaboration strategies (intermediate or late) based on spatial confidence maps.
Extensive evaluations on the V2X-R, V2X-Radar and UAV3D-mini datasets demonstrate that \ourmethod achieves the best overall trade-off between accuracy and communication cost among existing collaborative-perception methods.
Relative to SOTA \textit{Where2comm}, \ourmethod uses only 41\% of the bandwidth on V2X-R and 26\% on V2X-Radar while improving performance by 0.78\% and 0.75\% respectively.
Our code and models are available at https://github.com/AICPS/HydraCollab.
\end{abstract}

%% file: iros_camera_ready_sections/1_iros_intro.tex
\section{INTRODUCTION}
Collaborative-perception enhances performance in multi-robot systems by enabling robots to communicate perceptual information, overcoming individual limitations such as occlusions and restricted sensor range, and thereby improving situational awareness \cite{han2023survey}. Accurate perception is critical to ensure the safety of robotic systems such as autonomous vehicles \cite{huang2025v2x, yang2024v2x-radar} and UAV swarms \cite{uav3d2024}, where perceptual failures can lead to catastrophic results including collisions and injuries. Often, robots may wish to exchange perceptual information with non-robotic sensing platforms, such as infrastructure-mounted sensors in autonomous driving scenarios. We therefore use the term ``agent" to refer to each participant in collaborative-perception.

\begin{figure}[!t]
\centering
\includegraphics[width=0.8\columnwidth]{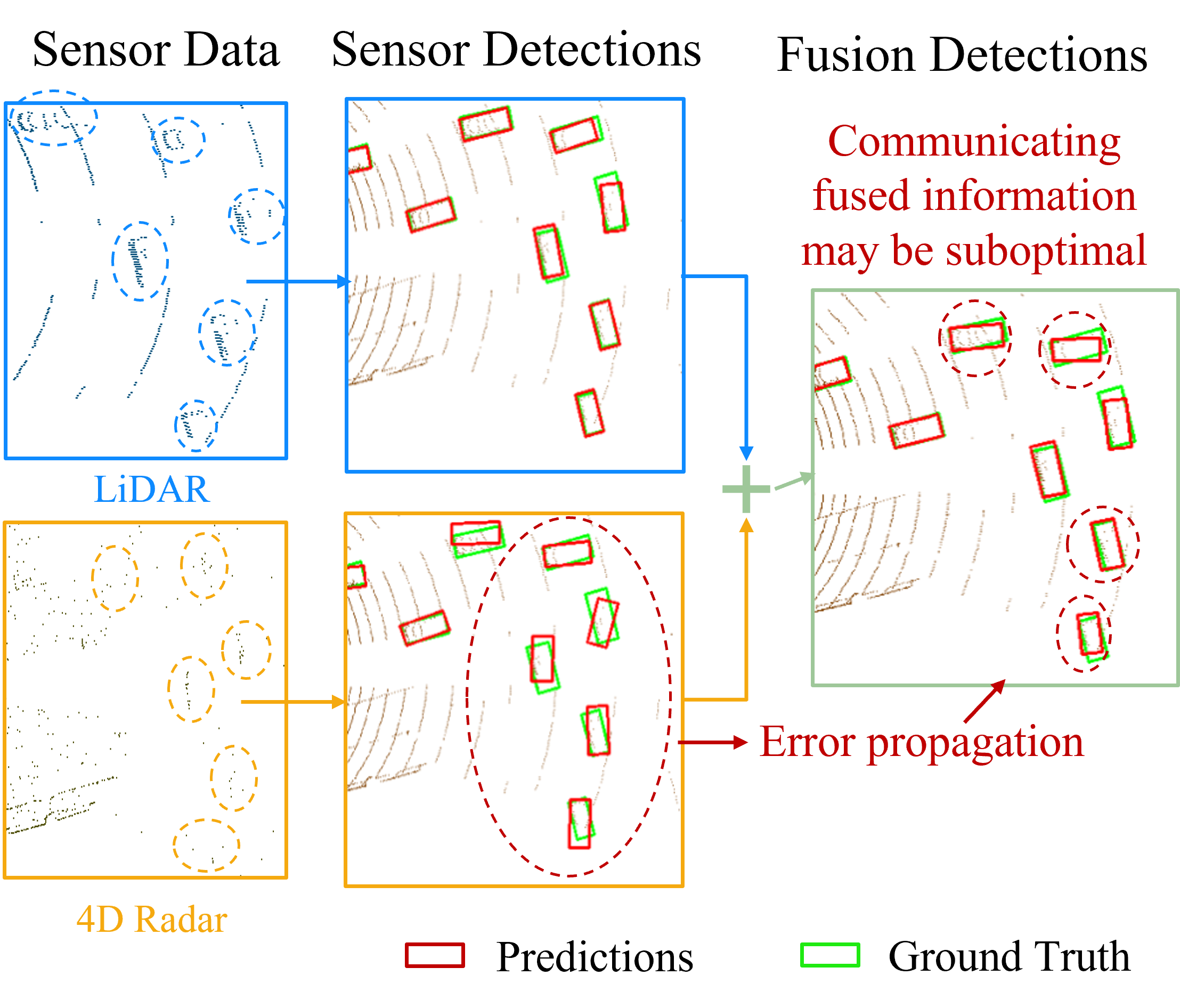}
\caption{LiDAR with noisy Radar fusion leads to suboptimal detections. Communicating only LiDAR information can improve both performance and communication bandwidth of the perception system (V2X-R dataset).}
\label{fig:motivation}
\end{figure}

In collaborative-perception, a fundamental trade-off exists between perception accuracy and communication bandwidth: sharing more information enhances performance but increases communication overhead. Given the limitations of real-world communication systems in supporting high-bandwidth exchange in real time, a key challenge is to maximize perception gains while minimizing communication costs. This trade-off has motivated three primary collaboration strategies, distinguished by the type of information shared. Late collaboration transmits only final detection results, minimizing bandwidth but limiting collaboration to perception outputs\cite{han2023survey, rauch2012car2x_latefusion,rawashdeh2018collaborative_latefusion}. Early collaboration exchanges raw sensor data, providing maximum information richness at the cost of high bandwidth requirements \cite{han2023survey, chen2019cooper_earlyfusion}. Intermediate collaboration seeks a balance by sharing encoded feature representations, making it the dominant paradigm in many recent systems \cite{hu2023collaboration, hu2022where2comm}, as it preserves informative content while significantly reducing communication overhead. Further advances reduce communication demands through selective strategies that determine what regions of interest to share \cite{hu2022where2comm}, which agents to communicate with \cite{liu2020who2com}, and when communication should occur \cite{liu2020when2com}.

Despite recent advances, existing collaborative-perception methods exhibit several key limitations that motivate this work. One limitation is that current approaches such as \cite{hu2022where2comm} and \cite{lu2024heal} typically assume that individual agents utilize all available sensors uniformly, regardless of environmental context or data quality. However, recent work in \cite{malawade2022hydrafusion} demonstrates that adaptive sensor selection based on contextual factors such as adverse weather conditions or occlusions can significantly improve single-agent perception performance. Figure~\ref{fig:motivation} illustrates how noisy local fusion can produce suboptimal features for collaboration. As shown in Figure~\ref{fig:motivation}, the noise in the 4D Radar detections propagates into the fused (LiDAR + Radar) detections, leading to fusion detections that are worse than the results of LiDAR alone. This motivates selective feature sharing in collaborative settings, which could improve perceptual accuracy while simultaneously minimizing communication bandwidth by choosing only the best features to share (e.g. LiDAR in Figure~\ref{fig:motivation}). 


Another limitation of many existing collaborative-perception methods is that they apply uniform intermediate collaboration strategies across all spatial regions. While \textit{Where2comm} \cite{hu2022where2comm} introduces spatial selectivity to identify critical areas for collaboration, it is constrained to exclusively use intermediate collaboration within those selected regions. However, in regions where agents’ perceptual information does not overlap, intermediate collaboration offers limited benefit. Instead of exchanging feature-level data in these areas, agents lacking confident perceptual information should rely on high-level perception results from peers that cover the region, thereby conserving bandwidth. This suggests that a hybrid approach combining intermediate collaboration in regions of mutual confidence with late collaboration in regions of single-agent confidence could achieve superior bandwidth efficiency with minimal performance degradation.

To address these research gaps, we propose \ourmethod, a novel adaptive collaborative-perception framework with the following key contributions:
\begin{enumerate}
    \item \textbf{Collaboration-aware sensor gating} that adaptively selects the most informative sensor features to communicate, enhancing the effectiveness of heterogeneous data while reducing communication bandwidth.
    \item \textbf{Spatially-aware collaboration strategy} that combines intermediate and late collaboration, guided by spatial confidence maps to apply rich feature sharing only in regions where both agents have confident and overlapping sensor information.
    \item \textbf{Natural extension to heterogeneous agents} with different sensing capabilities.
\end{enumerate}

Experimental evaluation on the V2X-R \cite{huang2025v2x}, V2X-Radar \cite{yang2024v2x-radar}, and UAV3D-mini \cite{uav3d2024} datasets demonstrate \ourmethod's applicability to real-world communication networks where strict bandwidth constraints must be satisfied without compromising accuracy.

%% file: iros_camera_ready_sections/2_iros_related.tex
\section{RELATED WORKS}

\subsection{Collaborative-Perception}
Collaborative-perception is a key advantage of multi-agent systems. By communicating sensing information, agents are able to improve their perception and address limitations such as occlusion and limited sensor range~\cite{hu2023collaboration,hu2022where2comm,xu2022cobevt,xu2022opv2v,rauch2012car2x_latefusion,rawashdeh2018collaborative_latefusion,chen2019cooper_earlyfusion, li2025comamba_iros, wang2025cocmt_iros,zhao2025coopre_iros}.

Collaboration methods for collaborative-perception can be categorized into early, intermediate, and late collaboration approaches based on what information is communicated~\cite{han2023survey}. 

Early collaboration~\cite{chen2019cooper_earlyfusion} allows agents to share sensor data, achieving high detection accuracy but requires the largest communication bandwidth. Late collaboration allows agents to transmit their own detection results, resulting in the lowest communication cost but making the system less robust and more prone to local errors such as sensor noise and occlusions~\cite{rauch2012car2x_latefusion,rawashdeh2018collaborative_latefusion}.

Therefore, intermediate collaboration~\cite{hu2023collaboration, hu2022where2comm,li2021disconet,liu2020when2com,liu2020who2com,qiao2023adafusion,qu2024sicp,xu2022cobevt,xu2022opv2v,xu2022v2xvit,lu2024heal, li2025comamba_iros, wang2025cocmt_iros,zhao2025coopre_iros}, where agents share encoded feature representations, has been the dominant choice given its balance between performance and communication efficiency. 

To further reduce communication bandwidth, many works propose selective communication strategies. Who2com~\cite{liu2020who2com} introduces a handshake communication mechanism to select two agents for collaborative-perception. When2com~\cite{liu2020when2com} proposes a communication group with an attention mechanism that enables agents to learn when to communicate and which agents to include. Where2comm~\cite{hu2022where2comm} adopts spatial confidence maps to identify critical regions and transmit region-level features to improve communication efficiency. DiscoNet~\cite{li2021disconet} introduces an early collaboration model as a teacher for knowledge distillation and guides an intermediate collaboration model during training.

In addition to intermediate feature exchange methods, Liu et al.~\cite{liu2023region} proposed a hybrid framework that applies intermediate collaboration in overlapping regions and late collaboration elsewhere. However, their proximity-based partitioning relies solely on geometric overlap and ignores feature reliability. Consequently, unreliable or noisy features are still transmitted, wasting bandwidth without improving performance. We propose a collaborative-perception framework that uses spatial confidence maps to selectively transmit only reliable features, combining the efficiency of late collaboration with the robustness of intermediate collaboration.

\subsection{Adaptive Sensor Fusion}
Sensor fusion~\cite{huang2025l4dr,malawade2022hydrafusion, chen2025hyperdimensional, yu2025vikienet_sensor_attn,wang2022interfusion} has been extensively studied to leverage the complementary advantages of multiple sensors. Prior methods have mainly focused on local sensor fusion within a single agent, where the fusion of Camera, LiDAR, and Radar improves robustness under adverse conditions and enhances detection accuracy. Previous adaptive feature-level sensor fusion strategies typically apply weighting mechanisms~\cite{huang2025l4dr,malawade2022hydrafusion,chen2025hyperdimensional} or attention modules~\cite{yu2025vikienet_sensor_attn,malawade2022hydrafusion,wang2022interfusion} to dynamically adjust the contribution of each information source depending on the sensing context.

However, existing multi-sensor collaborative-perception frameworks assume agents perform full local sensor fusion before transmitting intermediate fusion features [2], [24], [33]. This design has several limitations: (1) it ignores the global context available through collaboration, which provides superior information for assessing sensor value; (2) it allows misalignment between collaborators' self-assessment and ego needs (e.g., a collaborator may undervalue a sensor that the ego critically requires); (3) it can waste bandwidth transmitting redundant or detrimental fusion features; (4) and it often assumes agents have homogeneous sensing capabilities. To address these challenges, we propose an adaptive collaborative-perception method that leverages spatial confidence maps to evaluate global context and select the most valuable features before collaboration, reducing communication overhead while maintaining robustness and supporting heterogeneous settings.

%% file: iros_camera_ready_sections/3_iros_problem.tex
\begin{figure*}[ht]
    \centering
    \includegraphics[width=0.8\textwidth]{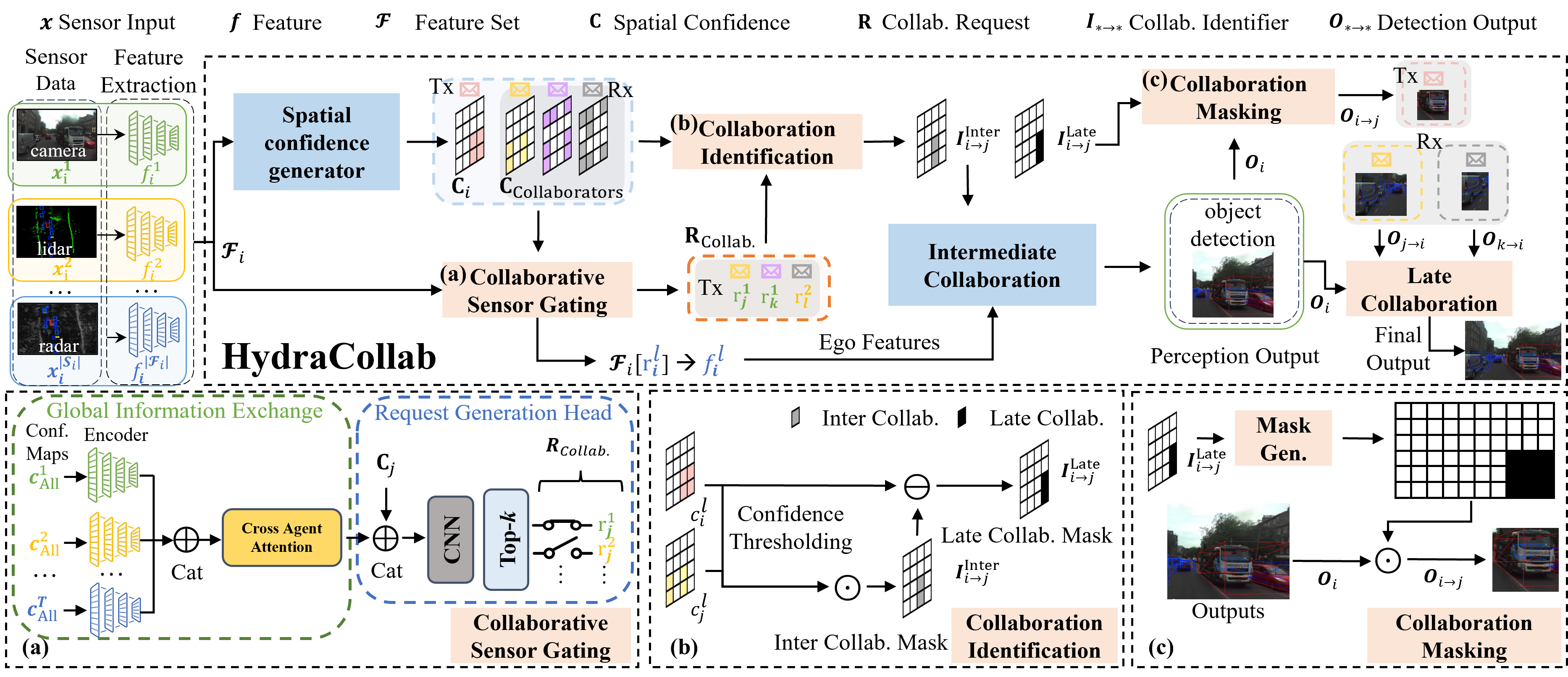}
    \caption{\ourmethod leverages spatial confidence maps~\cite{hu2022where2comm} to identify (a) the best sensor features to request from collaborators (Collaborative Sensor Gating), and (b) the collaboration strategy based on overlap/non-overlap regions for intermediate/late collaboration respectively (Collaboration Identification).}
    \label{fig:hydracollab}
\end{figure*}

\section{Problem Formulation}
We consider collaborative object detection in a multi-agent system. The objective is to select the most informative sensor features for collaboration and the best collaboration strategy for each spatial region (intermediate or late) to maximize perception accuracy and reduce communication overhead.

\subsection{Adaptive Collaborative-Perception Objective}
Formally, let $N$ denote the number of agents and $\mathcal{S}$ the global sensor set.
For each agent $i$, we define its available sensor subset $S_{i} \subseteq \mathcal{S}$ and the corresponding observations $X_i = \{x^{s}_{i}|s\in S_{i}\}$, where $x^{s}_{i}$ is the measurement from sensor $s$ of agent $i$.
For each observation $X_i$, we extract a feature set $\mathcal{F}_{i} = \Phi_{enc}(X_{i})$ through the encoder $\Phi_{enc}$ which can include individual sensor features and/or combinations of local fusion features similar to~\cite{malawade2022hydrafusion}. Consequently, the number of features $|\mathcal{F}_i|$ does not necessarily equal the number of sensors $|S_i|$.
The perception model $\mathcal{P}_{\theta}(\mathcal{F}_i, \{\mathcal{M}_{j\rightarrow i}\}_{j=1}^{N})$ produces object detections using both the local features $\mathcal{F}_i$, and the communicated messages from each collaborator $\{\mathcal{M}_{j\rightarrow i}\}_{j=1}^N$ where $\mathcal{M}_{j\rightarrow i}$ is the message transmitted from agent $j$ to agent $i$.
The objective is to maximize the perception performance of the ego agent $i$ while adhering to a communication budget $B$ by jointly optimizing the perception parameters $\theta$, the feature encoder $\Phi_{enc}$, and the inter-agent messages $\{\mathcal{M}_{j\rightarrow i}\}_{j=1}^{N}$:

\begin{equation}\label{perc_model}
\max_{\theta, \Phi_{enc}, \{\mathcal{M}_{j\rightarrow i}\}_{j=1}^N}{ \tau \big( \mathcal{P}_{\theta}(\Phi_{enc}(X_{i}), \{\mathcal{M}_{j\rightarrow i}\}_{j=1}^{N}), Y_{i} \big)},
\end{equation}

\begin{equation}\label{comm_budget}
~\text{s.t.} \sum^{N}_{j=1}|\mathcal{M}_{j\rightarrow i}| \leq B
\end{equation}

where $\tau(\hat{Y}_i,Y_i)$ is the perception evaluation metric comparing the perception output $\hat{Y}_i$ to the ground truth label $Y_i$ associated with each observation $X_i$ of agent $i$.
$\mathcal{M}_{j\rightarrow i}$ is an adaptive collaboration message:

\begin{equation}\label{comm_model}
\mathcal{M}_{j\rightarrow i} = \mathcal{M}^{Inter}_{j\rightarrow i} \cup \ \mathcal{M}^{Late}_{j\rightarrow i}
\end{equation}
where $\mathcal{M}^{Inter}_{j\rightarrow i}$ is the intermediate collaboration message containing perception features and $\mathcal{M}^{Late}_{j\rightarrow i}$ is the late collaboration message containing perception outputs.

%% file: iros_camera_ready_sections/4_iros_method.tex
\section{Methodology}
We present \ourmethod, an adaptive collaborative-perception framework using spatial confidence maps for 1) collaborative sensor gating and 2) adaptive collaboration to improve communication bandwidth efficiency.
Figure~\ref{fig:hydracollab} shows the overall methodology.

\subsection{Spatial Confidence Maps}
\label{sec:SpatialConfMaps}
To perform collaborative sensor gating and adaptive collaboration, an effective communication medium is needed to convey the importance of each sensor's contribution in each spatial region. \textit{Where2comm}\cite{hu2022where2comm} proposed the spatial confidence map which highlights perceptually critical features (e.g. areas containing objects) while omitting background features to save bandwidth.
Similarly, we generate spatial confidence maps of each agent over their feature set as:
\begin{equation}\label{conf_map}
\textbf{C}_{i} = \Phi_{gen}(\mathcal{F}_{i})=\{c^{l}_{i}|f^l_i\in \mathcal{F}_{i}\} \in [0,1]^{|\mathcal{F}_{i}|\times H\times W}
\end{equation}
where $\Phi_{gen}$ is the detection decoder, $l$ denotes the feature type, $c^{l}_{i}$ is the spatial confidence map of feature $f^l_i$ for agent $i$, and $H$ and $W$ are the height and width dimensions.

\subsection{Collaborative Sensor Gating} \label{sec:CollabSensorGating}
Collaborative sensor gating (Figure~\ref{fig:hydracollab}a) leverages all agents’ spatial confidence maps to select the most informative features for collaboration. 
We adopt the Compression--Sharing--Attention (CSA) structure from Attentive Fusion~\cite{xu2022opv2v}.
Spatial confidence maps are first sorted by feature type (e.g. Radar, LiDAR, or LiDAR-Radar Fusion for V2X-R dataset). In Figure~\ref{fig:hydracollab}a, $c^{l}_{All}$ denotes all spatial confidence maps of feature type $l$ and $T$ denotes the total number of feature types. 
Feature-type-specific convolutional encoders are utilized to extract high-level spatial-confidence information. 
The encoded confidence maps are then concatenated before applying multi-head self-attention to capture cross-agent global information.
The ego agent then predicts collaboration requests using a request generation head. Specifically, for each collaborator $j$, it outputs a request $r^{l}_{j}$ indicating whether to request feature $l$.
To compute $r^{l}_{j}$, the cross-agent global information is upsampled to the original map resolution via bilinear interpolation, concatenated with agent $j$'s confidence maps, and fed into a CNN-based decision network.
The network applies a softmax to output feature selection probabilities, followed by top-$k$ gating to produce the final set of requested features.

\subsection{Adaptive Collaboration}\label{sec:adapt_collab}
\subsubsection{Collaboration Identification}\label{sec:collabidentification}
Two agents will perform intermediate collaboration in regions with overlapping confidence, and late collaboration otherwise.
An intermediate collaboration mask is calculated through the dot product between their binary spatial confidence masks as 
$I^{Inter}_{i \rightarrow j} = \mathbf{1}(c^{l}_{i} \ge \lambda) \odot \mathbf{1}(c^{l}_{j} \ge \lambda) \in \{0, 1\}^{H \times W},$
where $\mathbf{1}$ is the indicator function evaluating to 1 when values are greater than or equal to the confidence threshold $\lambda$ and 0 otherwise.  
The late collaboration mask $I^{Late}_{i\rightarrow j}$ is computed by taking the difference between the intermediate collaboration mask $I^{Inter}_{i\rightarrow j}$ and the agent $i$'s confidence mask $\mathbf{1}(c^{l}_{i} \ge \lambda) $. Finally, a Gaussian filter is applied to smooth boundary regions in both $I^{Inter}_{i\rightarrow j}$ and $I^{Late}_{i\rightarrow j}$.

\subsubsection{Intermediate Collaboration}
During the intermediate collaboration stage, ego features $f^{l}_{i}$ are augmented by $I^{Inter}_{i \rightarrow j}$ through $f^{l}_{i \rightarrow j} = I_{i \rightarrow j}^{Inter} \odot f^{l}_{i}$ to transmit only the spatially correlated features from $i$ to $j$.
Subsequently, the ego agent $i$ receives the features from its collaborators that were requested after the gate in the same way messages are exchanged in \cite{hu2022where2comm}.
Finally, the ego features along with the collaborated intermediate features are fused together.
\subsubsection{Collaboration Masking}
In the late collaboration stage, as illustrated in Figure~\ref{fig:hydracollab} (c), agent $i$ first produces its own perception outputs $O_{i}$ through intermediate collaboration.  
The late collaboration mask \( I^{Late}_{i \rightarrow j} \) is upsampled and applied to these outputs to determine which detections should be transmitted to agent $j$ as  
$
O_{i \rightarrow j} = I^{Late}_{i \rightarrow j} \odot O_{i}
$.
From agent $i$'s perspective, after receiving $O_{j \rightarrow i} $, agent $i$ combines it with its own perception outputs $O_{i}$ to produce the final detection.

%% file: iros_camera_ready_sections/5_iros_exp.tex
\section{EXPERIMENTS}
\subsection{Datasets}
We evaluate \ourmethod on three multi-agent collaborative-perception datasets: V2X-R~\cite{huang2025v2x}, V2X-Radar~\cite{yang2024v2x-radar}, and UAV3D-mini~\cite{uav3d2024}. \textbf{V2X-R}~\cite{huang2025v2x} is a simulated benchmark built on OpenCDA~\cite{xu2021opencda} and CARLA~\cite{Dosovitskiy17carla}, providing LiDAR and 4D Radar data from scenes involving multiple Connected Autonomous Vehicles (CAVs) and infrastructure-mounted sensors. Each scene includes up to five collaborating agents that share complementary viewpoints for 3D detection. To ensure \ourmethod adapts to real-world data, we further benchmark on \textbf{V2X-Radar}~\cite{yang2024v2x-radar}, the first real-world collaborative-perception dataset that includes both LiDAR and 4D Radar from a single vehicle and an infrastructure-mounted sensor. Finally, to demonstrate applicability to diverse embodiments and settings, we evaluate on \textbf{UAV3D-mini}~\cite{uav3d2024}, a 10K-image subset of the simulated UAV3D dataset~\cite{uav3d2024}, co-simulated by CARLA~\cite{Dosovitskiy17carla} and AirSim~\cite{shah2017airsim}, containing data from five UAVs, each equipped with five cameras (front, rear, left, right, and bottom views). Drones fly in a cross-shaped formation to capture complementary and partially overlapping views of 3D environments.

\begin{figure*}[ht]
    \centering
    \includegraphics[width=0.8\textwidth]{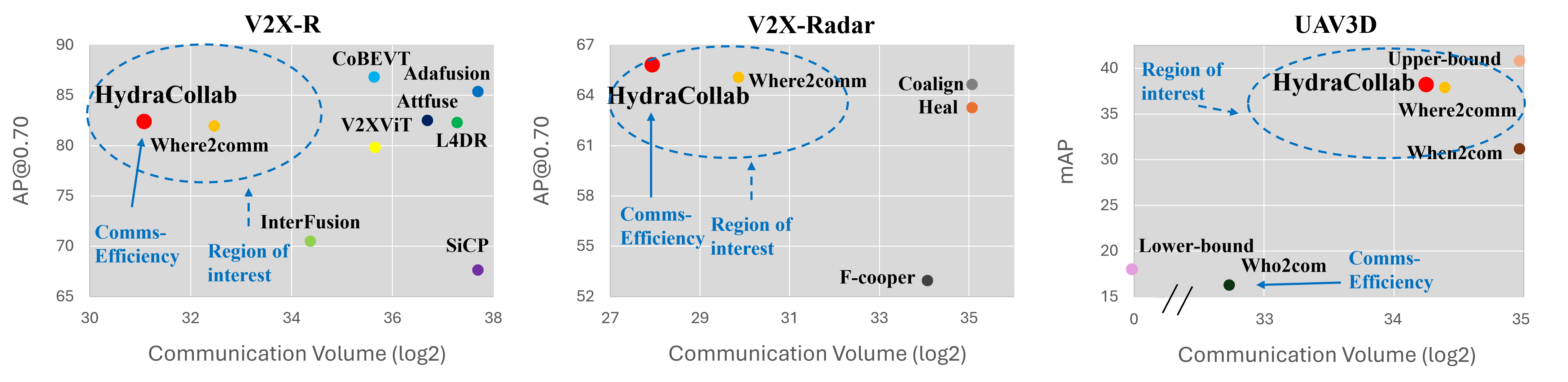}
    \caption{Performance-Communication Trade-off. \textnormal{\ourmethod} consistently achieves superior accuracy-bandwidth trade-off compared to SOTA methods.}
    \label{fig:hydracollab_alltradeoff}
\end{figure*}
\noindent

\subsection{Training}\label{sec:gate_training}
\ourmethod is trained end-to-end with a soft gating mechanism, which is converted to top-$k$ selection during inference.
Since different features may incur varying communication costs, we optimize the gating policy using a bandwidth-regularized loss:

\begin{equation}
\mathcal{L}_{G_\pi}
= \mathcal{L}_{\text{perception}}
+ \gamma \sum_{j} |f_j^l| ,
\end{equation}

where $\mathcal{L}_{\text{perception}}$ is the detection loss, $|f_j^l|$ is the number of bytes required to transmit feature $f_j^l$ from agent $j$ to ego, and $\gamma$ is a trade-off parameter.

\subsection{Quantitative Evaluation} 
\subsubsection{V2X collaboration}
For the V2X-R \cite{huang2025v2x} and V2X-Radar \cite{yang2024v2x-radar} datasets, the objective is to detect accurate 3D bounding boxes of vehicles in complex traffic scenes. We adopt Average Precisions (AP) under the Intersection over Union (IoU) threshold of 0.7 as the evaluation metric. Feature gating is applied over LiDAR, Radar, and fused LiDAR+Radar features with the top $k=1$ feature selected.

\subsubsection{Multi-Drone Collaboration}
The second task focuses on collaborative-perception in a UAV swarm setting. Our objective is multi-drone collaborative 3D object perception from a bird’s-eye view perspective.
We adopt mean Average Precision (mAP) under distance-based matching criteria as the evaluation metric as defined in \cite{uav3d2024}. Each camera is considered as an independent information source and sensor gating is applied for each perspective with the top $k=4$ cameras selected.

\subsubsection{Communication volume}
To measure the communication volume of intermediate collaboration we follow the same formulation as Where2comm~\cite{hu2022where2comm} where the pre-round confidence maps exchange were omitted as they are negligible compared to the sensor features (less than 1\%).
Let $F_j^G \subseteq \mathcal{F}_j$ denote the set of features requested by the gate (Section \ref{sec:CollabSensorGating}) from agent $j$ to agent $i$.
For each feature $f_j^l \in F_j^G$, a sparse feature $f^{l}_{j \rightarrow i} = I_{j \rightarrow i}^{Inter} \odot f^{l}_{j} \in \mathbb{R}^{H \times W \times C}$, is communicated 
where $I_{j \rightarrow i}^{Inter}$ is the binary intermediate collaboration mask extracted from Section~\ref{sec:adapt_collab}.
The message transmitted in bytes for each feature is defined as:
\begin{equation}
|\mathcal{M}^{Inter}_{j \rightarrow i}\big(f^{l}_{j \rightarrow i}\big)|
= |I_{j \rightarrow i}^{Inter}| \times D_j^l \times 32/8,
\end{equation}
where $|I_{j \rightarrow i}^{Inter}|$ denotes the number of selected grids, $D_j^l$ is the channel dimension of feature $f_j^l$, and $32$ represents float32 data type, and division by $8$ converts bits into bytes.

Late collaboration message transmission is defined as:
\begin{equation}
|\mathcal{M}^{Late}_{j \rightarrow i}\!\big(O_{j \rightarrow i}\big)|
= |O_{j \rightarrow i}| \times D_b \times 32/8,
\end{equation}
where $|O_{j \rightarrow i}|$ denotes the number of transmitted detections and $D_b$ represents the feature dimension of each detected object. Each detected object is encoded by its geometric parameters and confidence score, and $32$ represents  float32 data type, and division by $8$ converts bits into bytes.

Total transmitted message is defined as:
\begin{equation}
|\mathcal{M}_{j \rightarrow i}|
= 
\sum_{f^{l}_{j} \in F_j^G}
|\mathcal{M}^{Inter}_{j \rightarrow i}\!\big(f^{l}_{j \rightarrow i}\big)|
+ |\mathcal{M}^{Late}_{j \rightarrow i}\!\big(O_{j \rightarrow i}\big)|
,
\end{equation}
where $| \mathcal{M}_{j \rightarrow i}|$ denotes the total communication volume from agent $j$ to agent $i$, and $F_j^G$ is the set of all features selected by the gating module.

\begin{table}[ht]
\caption{Performance and communication comparison on V2X-R.}
\centering
\begin{tabular}{lcc}
\toprule
\textbf{Method} & \textbf{AP@0.7} & \textbf{Comm (log2)} \\
\midrule
Adafusion~\cite{qiao2023adafusion}    & 85.37 & 37.69 \\
CoBEVT~\cite{xu2022cobevt}       & 86.82 & 35.63 \\
Attfuse~\cite{xu2022opv2v}      & 82.51 & 36.69 \\
V2XViT~\cite{xu2022v2xvit}       & 79.83 & 35.65 \\
L4DR~\cite{huang2025l4dr}         & 82.31 & 37.28 \\
InterFusion~\cite{wang2022interfusion}  & 70.51 & 34.37 \\
SiCP~\cite{qu2024sicp}         & 67.63 & 37.69 \\
Where2comm~\cite{hu2022where2comm}   & 81.96 & 32.47 \\
\hline
\textbf{\ourmethod}   & \textbf{82.74} & \textbf{31.18} \\
\bottomrule
\end{tabular}
\label{tab:v2xr_table}
\end{table}

\subsubsection{Performance and Communication Results}
Across all three datasets, \ourmethod achieves Pareto-optimal performance, outperforming all methods with similar or lower bandwidth while approaching the accuracy of methods requiring much higher communication costs. 

\textbf{V2X-R}~\cite{huang2025v2x}.
On the V2X-R dataset, we compare our proposed \ourmethod with prior baselines on the trade-off between performance (AP@0.7) and communication bandwidth, as shown in Table~\ref{tab:v2xr_table} and Figure~\ref{fig:hydracollab_alltradeoff}.
Among all collaborative-perception models, \ourmethod uses the lowest communication bandwidth while outperforming Where2comm~\cite{hu2022where2comm}, V2XViT~\cite{xu2022v2xvit}, L4DR~\cite{huang2025l4dr}, InterFusion~\cite{wang2022interfusion}, SiCP~\cite{qu2024sicp}, and Attfuse~\cite{xu2022opv2v}.
Building on the communication-efficient SOTA Where2comm~\cite{hu2022where2comm}, \ourmethod further improves performance by 0.78\% while requiring only 41\% of communication bandwidth of Where2comm.
In contrast, Adafusion~\cite{qiao2023adafusion} and CoBEVT~\cite{xu2022cobevt} are performance-oriented models that achieve higher accuracy but at the cost of $91\times$ and $22\times$ more communication bandwidth than \ourmethod, respectively.

\begin{table}[ht]
\caption{Performance and communication comparison on V2X-Radar.}
\centering
\begin{tabular}{lcc}
\toprule
\textbf{Method} & \textbf{AP@0.7} & \textbf{Comm (log2)} \\
\midrule
F-Cooper~\cite{chen2019fcooper}      & 52.95 & 34.07\\
CoAlign~\cite{lu2023coalign}       & 64.66 & 35.07\\
HEAL~\cite{lu2024heal}          & 63.27 & 35.07\\
Where2comm~\cite{hu2022where2comm}    & 65.07 & 29.86\\
\hline
\textbf{\ourmethod}   & \textbf{65.82} & \textbf{27.94} \\
\bottomrule
\end{tabular}
\label{tab:v2xRadar_table}
\end{table}

\textbf{V2X-Radar}~\cite{yang2024v2x-radar}. On the real-world V2X-Radar dataset, \ourmethod outperforms all evaluated baselines. Since V2X-Radar does not provide an official LiDAR--Radar fusion benchmark, we follow the multi-sensor fusion pipeline from V2X-R~\cite{huang2025v2x} to implement both \ourmethod and the evaluated baselines for a fair and consistent comparison. As shown in Table~\ref{tab:v2xRadar_table} and Figure \ref{fig:hydracollab_alltradeoff}, \ourmethod achieves the highest detection accuracy while using the lowest communication overhead among all evaluated methods, outperforming F-Cooper~\cite{chen2019fcooper}, CoAlign~\cite{lu2023coalign}, HEAL~\cite{lu2024heal}, and Where2comm~\cite{hu2022where2comm}. Compared to Where2comm~\cite{hu2022where2comm},
\ourmethod improves performance by 0.75\% while requiring only 26.4\% of the communication bandwidth of Where2comm.

\begin{table}[ht]
\centering
\caption{Performance and communication comparison on UAV3D-mini.}
\begin{tabular}{lcc}
\toprule
\textbf{Method} & \textbf{mAP} & \textbf{Comm (log2)} \\
\midrule
Lowerbound   & 18.0  & 0.00  \\
Upperbound   & 40.8  & 34.97 \\
Who2com\cite{liu2020who2com}      & 16.3  & 32.73 \\
When2com\cite{liu2020when2com}     & 31.2  & 34.97 \\
Where2comm\cite{hu2022where2comm}   & 37.9  & 34.39 \\
\hline
\textbf{\ourmethod}   & \textbf{39.1}  & \textbf{34.21} \\
\bottomrule
\end{tabular}
\label{tab:uav3d_table}
\end{table}

\textbf{UAV3D-mini}~\cite{uav3d2024}. 
In the multi-drone collaborative-perception task, we compare \ourmethod against existing methods in Table~\ref{tab:uav3d_table} and Figure~\ref{fig:hydracollab_alltradeoff}. Among intermediate collaboration approaches, \ourmethod achieves the highest accuracy, outperforming Where2comm~\cite{hu2022where2comm}, When2com~\cite{liu2020when2com}, and Who2com~\cite{liu2020who2com}. We additionally report a theoretical Upperbound that fuses all agents’ raw features with maximal communication, and a Lowerbound that uses only single-agent detections without inter-agent communication. Overall, \ourmethod improves over Where2comm~\cite{hu2022where2comm} by $3.17\%$ while using only $88\%$ of the communication bandwidth. 

\begin{figure*}[!ht]
    \centering
    \includegraphics[width=0.75\textwidth]{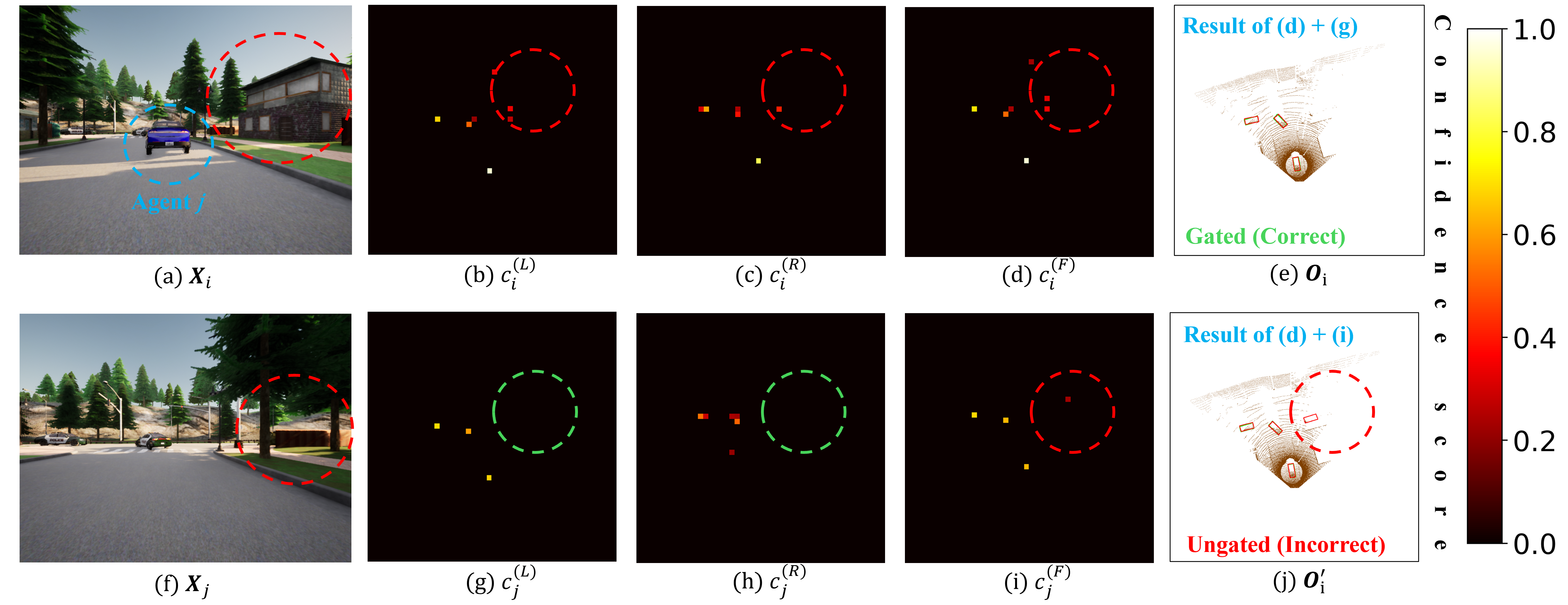}
    \caption{Example of top-1 Collaborative Sensor Gating on V2X-R dataset. For agents $i$ and $j$, their individual spatial confidence maps are shown using LiDAR ($c^{L}_{i,j}$), Radar ($c^{R}_{i,j}$) and L+R fusion ($c^{F}_{i,j}$). Fusing the L+R features from both agents amplifies noise, resulting in a false positive in (j) (red circle). Collaborative sensor gating in \textnormal{\ourmethod} selects fused features from agent $i$ and LiDAR features from agent $j$ to produce the correct result (e).}
    \label{fig:qualitative_inter_fusion_v2xr}
\end{figure*}
\begin{figure*}[!ht]
    \centering
    \includegraphics[width=0.75\textwidth]{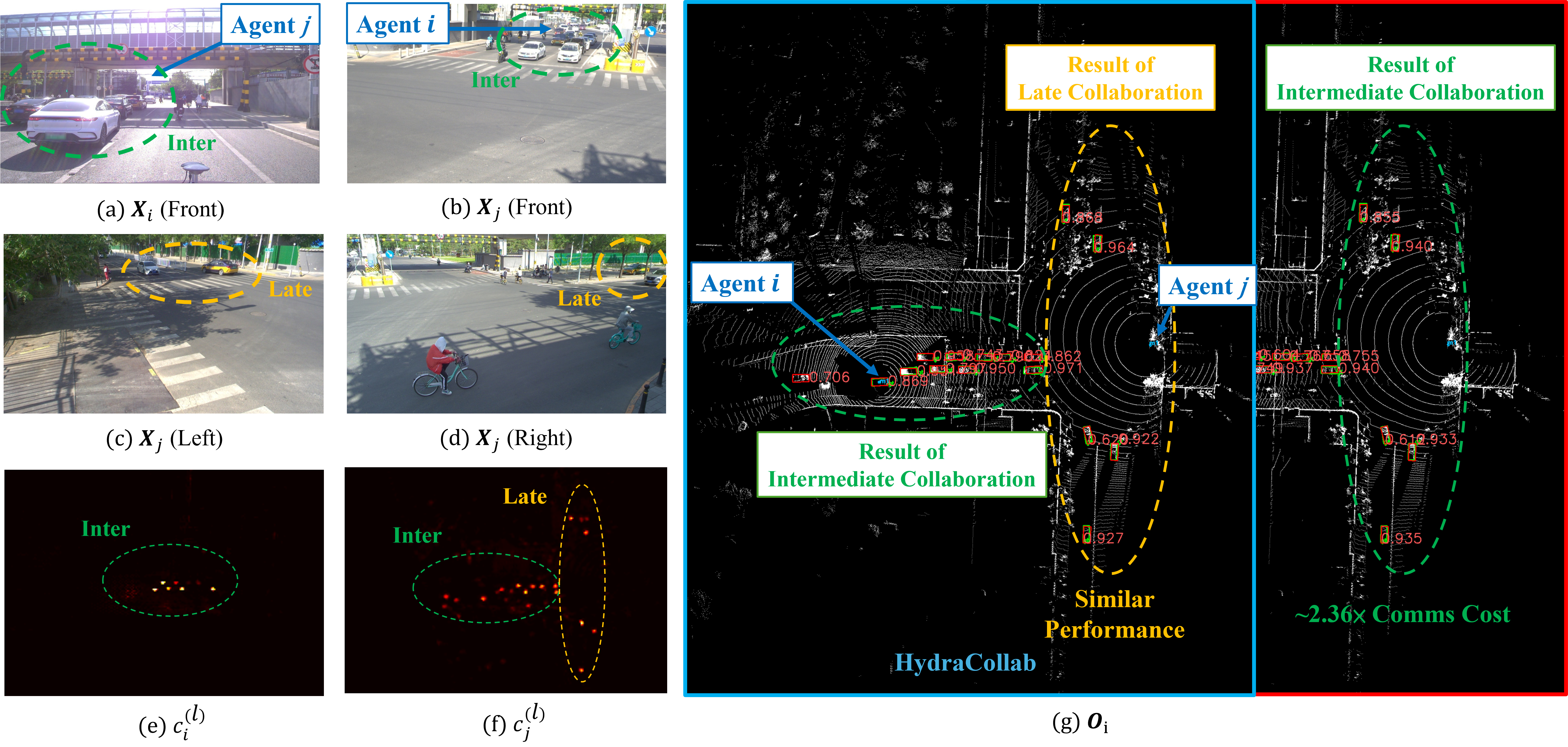}
    \caption{Visualization of adaptive collaboration at an intersection in V2X-Radar with one vehicle agent $i$ and one infrastructure agent $j$. (a) Front view of agent $i$. (b--d) Front, left, and right views of agent $j$. (e--f) Spatial confidence maps for agents $i$ and $j$, where the green circle marks a region jointly observed with high confidence, and the yellow circle in (f) marks a region confidently observed only by agent $j$ (out of view for agent $i$). Accordingly, \textnormal{\ourmethod} performs intermediate collaboration in the green region and late collaboration in the yellow region producing the detection results in (g).}
    \label{fig:qualitative_late_fusion}
\end{figure*}
\subsection{Qualitative Analysis}
\subsubsection{Visualization of Collaborative Sensor Gating on V2X-R} Figure~\ref{fig:qualitative_inter_fusion_v2xr} illustrates how Collaborative Sensor Gating enhances the performance of \ourmethod by choosing the best sensor features to collaborate. In this scenario, agent~$i$ (ego) and agent~$j$ (collaborator) are driving back-to-front, and both have obstacles on their right-hand side, as marked by the red circles in Figure~\ref{fig:qualitative_inter_fusion_v2xr} (a, f).
Figure~\ref{fig:qualitative_inter_fusion_v2xr} (b–d and g-i) display the spatial confidence maps of agents $i$ and $j$ under LiDAR (L), Radar (R), and L+R (F) fusion sensor selections. Without collaborative sensor gating, both agents would select fusion features to communicate. However, buildings and walls introduce noise that creates false high-confidence anchors in the circled regions. Consequently, using the fused features from all agents leads to the false positive detection circled in Figure~\ref{fig:qualitative_inter_fusion_v2xr} (j). In contrast, the collaborative sensor gating mechanism will evaluate all 6 spatial confidence maps and select the less noisy LiDAR features from agent $j$ and fusion features from agent $i$ to produce the correct result in  Figure~\ref{fig:qualitative_inter_fusion_v2xr} (e) while simultaneously reducing communication bandwidth by not transmitting Radar data.

\subsubsection{Visualization of Adaptive Collaboration on V2X-Radar}
Figure~\ref{fig:qualitative_late_fusion} visualizes the behavior of Adaptive Collaboration in \ourmethod. The ego agent $i$ is a vehicle approaching an intersection, and the collaborator $j$ is an infrastructure node. Due to the limited visibility of agent $i$, the yellow region in Figure~\ref{fig:qualitative_late_fusion}(f) is only confidently observed by agent $j$, and \ourmethod therefore uses late collaboration. In the mutually confident green region, \ourmethod applies intermediate collaboration to exploit shared visibility. \ourmethod produces accurate predictions in both regions while reducing communication by $2.36\times$ in this specific frame.

\subsection{Heterogeneous Agents}
We evaluate the robustness of \ourmethod under heterogeneous sensing capabilities to better reflect practical deployments where agents may not carry identical sensor suites. Using V2X-R~\cite{huang2025v2x}, we create a heterogeneous setting by restricting one agent to LiDAR-only or Radar-only, while all other agents retain both sensors. Table~\ref{tab:hetero_results} reports detection performance for each restricted-agent setting. The largest performance drop occurs when CAV 1 (the ego vehicle) is restricted to Radar-only, as LiDAR is generally the more reliable modality. Nonetheless, most other heterogeneous configurations remain close to the homogeneous baseline, demonstrating the adaptability of \ourmethod.

\begin{table}[ht]
\centering
\caption{Heterogeneous collaboration results on V2X-R.}
\begin{tabular}{c|c|c}
\hline
\textbf{Agent} & \textbf{Sensor} & \textbf{AP@0.7} \\
\hline
Homogeneous & L+R & 82.74 \\
\hline
\multirow{2}{*}{CAV 1} 
 & LiDAR only & 80.64 \\
 & Radar only & 64.01 \\
\hline
\multirow{2}{*}{CAV 2} 
 & LiDAR only & 81.90 \\
 & Radar only & 71.95 \\
\hline
\multirow{2}{*}{CAV 3} 
 & LiDAR only & 82.69 \\
 & Radar only & 81.96 \\
\hline
\multirow{2}{*}{CAV 4} 
 & LiDAR only & 82.61 \\
 & Radar only & 81.92 \\
\hline
\multirow{2}{*}{Infrastructure} 
 & LiDAR only & 82.61 \\
 & Radar only & 81.59 \\
\hline
\end{tabular}
\label{tab:hetero_results}
\end{table}

\subsection{HydraCollab Ablation}
\begin{table}[!ht]
\caption{Ablation study of \textnormal{\ourmethod} on V2X-R Dataset.}
\resizebox{0.45\textwidth}{!}{
\centering
\begin{tabular}{cc|cc}
\hline
\textbf{Collaborative} & \textbf{Adaptive} & \multirow{2}{*}{\textbf{AP@0.7}} & \multirow{2}{*}{\textbf{Comm (log2)}} \\
\textbf{Sensor Gating} & \textbf{Collaboration} & & \\
\hline
& & 81.96 & 32.47 \\
\checkmark & & 82.65 {\scriptsize (0.69 $\uparrow$)} & 31.84 {\scriptsize (35.4\% $\downarrow$)} \\
& \checkmark & 82.36 {\scriptsize (0.40 $\uparrow$)} & 31.39 {\scriptsize (52.7\% $\downarrow$)} \\
\checkmark & \checkmark & \textbf{82.74} {\scriptsize (0.78 $\uparrow$)} & \textbf{31.18} {\scriptsize (59.1\% $\downarrow$ )} \\
\hline
\textbf{Gate Type} & \textbf{Selection Rate (\%)} & \textbf{AP@0.7} & \textbf{Comm (log2)} \\
\hline
LiDAR & 1.00 / 0.00 / 0.00 & 79.36 & 30.08 \\
Radar & 0.00 / 1.00 / 0.00 & 4.02 & 30.85 \\
L+R & 0.00 / 0.00 / 1.00 & 82.32 & 31.38 \\
Random & 0.34 / 0.33 / 0.33 & 65.79 & 30.78 \\
\hline
\end{tabular}
}
\label{tab:ablation}
\end{table}
We conduct an ablation study on V2X-R~\cite{huang2025v2x} to assess the contribution of each \ourmethod component, as summarized in Table~\ref{tab:ablation} (top).
The baseline model is the vanilla intermediate collaboration with spatial confidence maps. Adding Collaborative Sensor Gating improves the baseline AP@0.7 performance to 82.65\%  and reduces the communication bandwidth by 35.4\%. This gain stems from (i) adaptively selecting the most suitable sensor subset for each frame and (ii) avoiding continuous transmission of all sensor features from every vehicle.
Adaptive Collaboration reduces the communication bandwidth by 52.7\% without degrading performance by 
employing late collaboration where the spatial confidence maps of different agents do not overlap.
Combining the two, \ourmethod lowers communication bandwidth by 59.1\% while increasing AP@0.7 by 0.78\% over the baseline. We observe a consistent trend where detection accuracy rises across settings even as communication bandwidth progressively decreases, indicating that the two components work together to reduce communication without compromising performance.
We further ablate the effectiveness of Collaborative Sensor Gating by comparing different gates.
From Table~\ref{tab:ablation} (bottom), we see that Collaborative Sensor Gating outperforms all fixed and random gates, empirically demonstrating our gates ability to select informative features for collaboration. 
For an in-depth study of context-aware sensor gating we refer the readers to~\cite{malawade2022hydrafusion}.

\begin{figure}[ht]
\centering
\includegraphics[width=0.75\columnwidth]{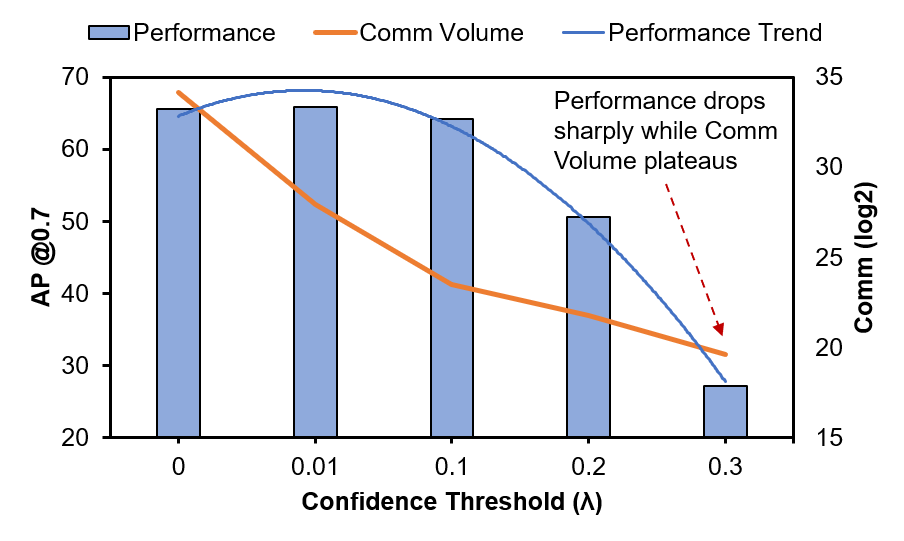}
\caption{Ablation study of $\lambda$ in Adaptive Collaboration on V2X-Radar}
\label{fig:sens}
\end{figure}
Figure~\ref{fig:sens} ablates the effects of the feature confidence threshold $\lambda$ used in Adaptive Collaboration (Section~\ref{sec:collabidentification})
to filter out noise before collaboration.
As $\lambda$ crosses 0.2, useful features are filtered out, resulting in performance drops.
On the other hand, communication volume drops as more features are filtered out but eventually plateaus.
$\lambda$ is tuned based on the sensitivity sweep as the last step before model deployment.

\subsection{Computation Complexity}
\begin{table}[ht]
\centering
\caption{Computational complexity comparison on V2X-Radar.}
\resizebox{\columnwidth}{!}{
\begin{tabular}{l|cc|cc}
\hline
\textbf{Model Components} 
& \multicolumn{2}{c|}{\textbf{\textnormal{\textbf{\ourmethod}}}} 
& \multicolumn{2}{c}{\textbf{Where2comm}~\cite{hu2022where2comm}} \\
\cline{2-5}
& \textbf{GFLOPs} & \textbf{Params (M)} 
& \textbf{GFLOPs} & \textbf{Params (M)} \\
\hline
Multi-Sensor Local Encoding 
& 658.89 & 1.09
& 659.10 & 1.09 \\
Spatial Conf. Map Generator 
& 3076.67$^{*}$ & SHARED 
& 1025.56 & SHARED \\
\textbf{Collaborative Sensor Gating} 
& \textbf{5.69} & \textbf{1.186} 
& -- & -- \\
Intermediate Collaboration 
& 559.42 & 13.65
& 559.42 & 13.65 \\
\textbf{Late Collaboration} 
& \textbf{319.44} & \textbf{SHARED} 
& -- & -- \\
\textbf{Collab. ID + Masking} 
& $\sim$\textbf{0.00} & $\sim$\textbf{0.00} 
& -- & -- \\
Downstream Task Heads 
& 387.89 & 0.01 
& 193.94 & 0.01 \\
\hline
\textbf{Total} 
& \textbf{5008.00} & \textbf{15.931} 
& \textbf{2438.02} & \textbf{14.75} \\
\hline
\end{tabular}
}
\label{tab:computation}
\end{table}

Table~\ref{tab:computation} presents the computational complexity of \ourmethod and Where2comm~\cite{hu2022where2comm}, measured on one frame with two agents on V2X-Radar.
Compared to~\cite{hu2022where2comm}, \ourmethod is a multi-sensor ensemble variant which naturally uses more FLOPs, primarily due to the per-sensor$^{*}$ Spatial Conf. Map generation.
Practically, these per-sensor maps are computed independently; when batch-parallelized, their cost manifests as throughput rather than critical-path latency.
The Collaborative Sensor Gating module uses a lightweight Global Information Exchange module and a CNN-based Request Generation Head, contributing only 0.11\% of the total FLOPs while accounting for 7.44\% of the total parameters.
Collaboration ID+Masking involves element-wise products, differences, and up-sampling, which are computationally negligible and add no parameters.
Overall, the proposed gating and masking components incur minimal overhead for substantial gains in performance and bandwidth efficiency.

%% file: iros_camera_ready_sections/6_iros_conclusion.tex
\section{CONCLUSION}
\ourmethod is an adaptive collaborative-perception architecture designed to significantly reduce communication bandwidth.
\ourmethod combines collaborative sensor gating, which adaptively transmits only the most informative sensor features, with spatially adaptive collaboration that dynamically chooses the best collaboration strategy (intermediate or late) for each spatial region.
Evaluated on both simulated and real-world datasets, \ourmethod demonstrates significant bandwidth savings while improving perception performance compared to existing methods.
While \ourmethod leverages Where2comm's~\cite{hu2022where2comm} spatial confidence maps as the necessary context features for decision making, future work aims to explore more effective alternatives that can further improve adaptive collaborative-perception systems.
In conclusion, \ourmethod's spatial-confidence driven, adaptive approach overcomes the rigid trade-offs of previous systems, paving the way for more scalable, efficient, and capable multi-agent collaborations in bandwidth-constrained applications.